\documentclass[10pt,twocolumn,letterpaper]{article}

\usepackage[letterpaper,top=0.72in,bottom=0.72in,left=0.72in,right=0.72in,columnsep=0.24in]{geometry}
\usepackage{fontspec}
\usepackage{microtype}
\usepackage{amsmath,amssymb,bm}
\usepackage{graphicx}
\usepackage{booktabs}
\usepackage{multirow}
\usepackage{array}
\usepackage{caption}
\usepackage{subcaption}
\usepackage{enumitem}
\usepackage[hidelinks]{hyperref}
\usepackage{url}
\graphicspath{{figure/}}
\setlist[itemize]{leftmargin=1.3em,noitemsep,topsep=2pt}

\newcommand{\ours}{\textbf{Ours}}
\newcommand{\Rres}{R_{\mathrm{res}}}

\title{\textbf{LowAux-RDNet: Low-Pass Residual Supervision\\
with Scene-Balanced Real-World Training\\
for Single-Image Reflection Removal}}
\author{Jizhong Li\\Fudan University}
\date{}

\begin{document}
\raggedbottom
\maketitle

\begin{abstract}
Single-image reflection removal aims to recover a clean transmission layer
from one image captured through glass. We study an explicit decomposition
pipeline built on RDNet and introduce LowAux, a training-only low-pass
reflection auxiliary objective. The original residual target remains the main
reflection supervision, while symmetrically filtered prediction and target
provide a stable low-frequency constraint. We further incorporate
scene-balanced real pairs from RRW to broaden real-scene coverage and improve
cross-dataset generalization. To avoid evaluation discrepancies caused by
model-specific resizing, padding, output quantization, and metric code, we
build a unified public benchmark over CEILNet, Real20, Postcard, Objects, and
Wild. Under the same evaluator, the proposed system obtains a five-dataset
macro average of 27.546 dB PSNR, 0.9220 SSIM, 0.9751 NCC, and 0.004760 LMSE,
achieving the highest macro-average PSNR, SSIM, and NCC and the lowest LMSE
among the compared public checkpoints and internal variants.
Per-dataset and qualitative analyses show that the main benefit is a more
balanced performance across diverse reflection distributions, while clear
semantic reflections in Postcard remain challenging.
\end{abstract}

\section{Introduction}

Images captured through glass or other reflective mediums contain a mixture of a desired transmission
layer and an undesired reflection layer. Undesired reflections can severely degrade the imaging quality
of the target scene, which will in turn affect downstream tasks such
as object detection~\cite{wan2020reflection}. Single-image reflection removal
(SIRR) commonly approximates the observed mixture as the superposition of a
transmission layer and a reflection layer, i.e., $M\approx T+R$. Separating
these layers from a single observation is nevertheless severely ill-posed:
both layers may contain natural edges, textures, and complete semantic
objects, and real imaging additionally introduces attenuation, blur,
saturation, misalignment, and other unmodeled residual effects. Reflection
removal therefore depends not only on network capacity, but also on how layer
decomposition is supervised and how closely training data represent real
capture conditions.

Direct transmission-regression methods such as ERRNet~\cite{wei2019errnet}
can be strong restoration baselines, but they do not expose the removed
reflection-related component as an explicit branch. This makes it difficult
to impose structured residual supervision directly. We therefore adopt
RDNet~\cite{zhao2025rdnet} as an explicit decomposition backbone, so that the
model produces both a transmission estimate and a reflection-related residual
estimate.

Within the RDNet framework, we construct a reflection residual pseudo-label
$\Rres=M-T$. This residual may contain high-frequency contamination introduced
by imperfect registration, exposure changes, and acquisition noise. On the
other hand, directly replacing the full residual with a low-pass residual may
discard information that is useful for detailed layer separation. Therefore,
we retain RDNet's original residual supervision as the main objective and
introduce an additional low-pass reflection-residual auxiliary loss to constrain
more stable low-frequency reflection structures. Furthermore, to address the
performance trade-offs across different test distributions, we incorporate
RRW real paired data during training and use scene-balanced sampling to reduce
the training bias caused by redundant adjacent video frames.

Our contributions are:
\begin{itemize}
    \item a low-pass reflection-residual auxiliary loss on RDNet's explicit
    residual branch, which improves residual supervision without changing the
    inference architecture;
    \item a scene-balanced sampling strategy for RRW real paired data, designed
    to reduce adjacent-frame redundancy and improve aggregate cross-dataset
    performance;
    \item a unified public evaluation protocol that applies consistent data
    pairing, floating-point metric computation, and equal-weight dataset macro
    averaging to all available comparison models.
\end{itemize}

\section{Related Work}

\paragraph{Simplified imaging model and supervision.}
Most SIRR methods start from the simplified formation model
$I\approx T+R$, where a mixed observation is approximated by a desired
transmission image and an undesired reflection image. This model provides a
convenient notation and enables synthetic paired data, but it is not a
complete description of real glass imaging. Wen et al.~\cite{Wen_2019_CVPR}
show that linear superposition is insufficient for realistic reflection
synthesis and introduce nonlinear alpha blending to better mimic real
captures. More recent work further studies the gap between synthetic
reflection labels and real residual labels. GFRRN~\cite{Chen_GFRRN_2026}
explicitly points out that the synthetic reflection layer and the real
$I-T$ residual pseudo-label are not fully consistent, motivating unified label
generation and frequency-aware supervision. These observations are directly
related to our setting: the residual pseudo-label is useful, but its
high-frequency content may include alignment, exposure, and acquisition
artifacts rather than physically clean reflection.

\paragraph{Direct transmission regression.}
Single-image reflection removal is ill-posed because one observation must
explain two visually plausible layers. Early approaches relied on handcrafted
priors such as gradient sparsity, layer smoothness, and ghosting cues. These
assumptions are useful for restricted scenes but often fail when reflections
contain sharp edges or semantic objects. Deep methods instead learn a mapping
from mixed observations to clean transmission using paired synthetic and real
data. CEILNet~\cite{fan2017ceilnet} introduced a cascaded design that first
predicts an edge map and then reconstructs the transmission image, explicitly
using structure to guide restoration. Perceptual-loss-based separation
~\cite{zhang2018perceptual} showed that feature-space supervision can preserve
visually important content better than pixel losses alone.

ERRNet~\cite{wei2019errnet} addresses two practical limitations of direct
regression: insufficient contextual reasoning and imperfectly aligned real
training pairs. It combines VGG hypercolumns, channel and spatial context, and
an alignment-invariant high-level feature loss. This permits real unaligned
pairs to supplement accurately registered synthetic data without imposing
misleading pixel correspondence. ERRNet is therefore a strong representative
of direct transmission regression.
However, because it predicts only the transmission layer, the removed
reflection residual is not explicitly exposed as a branch that can receive
structured supervision.

\paragraph{Explicit decomposition and structure-aware priors.}
Rather than predicting only transmission, explicit decomposition methods
model interactions between transmission and reflection. BDN
~\cite{yang2018bdn} performs bidirectional estimation so that each layer helps
explain the other. IBCLN~\cite{li2020ibcln} repeatedly refines both
components through a cascade, progressively correcting separation errors.
RAGNet~\cite{Li_RAGNet_2020} first estimates reflection-related information
and then uses it to guide transmission recovery, while location-aware SIRR
~\cite{Dong_2021_ICCV} predicts a reflection confidence map to focus
processing on reflection-dominant regions. These methods show that reflection
structure, location, and interaction cues are useful beyond plain
transmission regression.

Other works introduce stronger priors or model-driven optimization into the
network. YTMT~\cite{hu2021ytmt} exchanges information between two streams and
treats features as useful or interfering depending on their destination.
DSRNet~\cite{hu2023dsrnet} promotes component synergy through specialized
feature interactions, while DSIT~\cite{hu2024dsit} uses interactive
dual-stream transformers to capture longer-range semantic dependencies.
DURRNet~\cite{Huang_DURRNet_2024} and DExNet~\cite{Huang_DExNet_2025}
represent deep-unfolding directions that embed prior constraints, such as
exclusion or joint priors, into learnable networks. RDNet~\cite{zhao2025rdnet}
introduces reversible multi-column interactions, input-conditioned prompting,
and hierarchical decoding. Its decoder exposes multi-scale outputs for both
branches, making the removed component directly observable during training.
This property is central to our work: instead of changing RDNet's
architecture, we use its residual branch as an interface for complementary
raw-residual and low-frequency supervision.

\paragraph{Real-world data and unified evaluation.}
Synthetic mixtures provide exact layer targets but cannot fully reproduce
real glass, illumination, camera motion, saturation, and spatially varying
reflections. Real data are consequently important for both training and
evaluation. SIR$^2$~\cite{wan2017sir2} provides controlled real test subsets
covering objects, postcards, and wild scenes. RRW~\cite{zhu2024rrw} expands
real-world coverage using many frames from diverse captured scenes. Its video
structure also creates substantial adjacent-frame redundancy; our
scene-balanced sampler treats scenes, rather than individual frames, as the
training units.

Reported results across reflection-removal papers are often not directly
comparable. Studies may use different subset sizes, resize Real20 differently,
quantize predictions before scoring, or implement PSNR and SSIM with different
conventions. Large subsets can also dominate an image-level average. We
therefore re-evaluate all available checkpoints with one floating-point
pipeline and report an equal-weight dataset macro average. This protocol
removes test-time implementation discrepancies, although differently trained
public checkpoints remain a system-level rather than strictly controlled
architectural comparison.

\section{Method}

\subsection{RDNet-Based LowAux}

Let $M\in[0,1]^{H\times W\times3}$ denote the mixed observation. We use the
simplified formation model
\begin{equation}
    M \approx T + R,
    \label{eq:formation}
\end{equation}
where $T$ and $R$ denote the ideal transmission and physical reflection
layers. Real captures do not exactly satisfy Eq.~\eqref{eq:formation}; a more
realistic abstraction can be written as
\begin{equation}
    M = T + R + \Phi(T,R),
    \label{eq:formation-expanded}
\end{equation}
where $\Phi(T,R)$ summarizes attenuation, nonlinear mixing, saturation,
misalignment, sensor noise, and other unmodeled acquisition residuals. In this
paper, Eq.~\eqref{eq:formation} is used as a notation convention, while the
actual reflection-branch target is the residual pseudo-label
\begin{equation}
    \Rres = M-T
    \label{eq:residual-pseudo-label}
\end{equation}
rather than the physically exact layer $R$.

We adopt RDNet~\cite{zhao2025rdnet} as the explicit decomposition backbone.
This exposes a residual-branch prediction that can be compared with
$\Rres$, allowing separate constraints on full residual detail and stable
low-frequency structure.

RDNet contains a pretrained hierarchy extractor (PHE), a
transmission-rate-aware prompt generator (TAPG), a multi-column reversible
encoder, and a hierarchy decoder. PHE supplies hierarchical visual features,
TAPG produces an input-conditioned separation prompt, and the reversible
encoder exchanges fine detail and broad context across columns and scales.
The encoder produces features rather than image-space layers. The hierarchy
decoder aggregates them and predicts multi-scale transmission and reflection:
\begin{equation}
 [\hat T_k,\hat R_k]=[M_k,M_k]-D_k(\mathcal{F}_k),
 \label{eq:rdnet-output}
\end{equation}
where $\mathcal{F}_k$ and $M_k$ denote hierarchical features and the resized
mixed input at scale $k$. In the following, $\hat R_k$ denotes the predicted
residual-branch output supervised by the residual pseudo-label
$R_{\mathrm{res},k}$, rather than a physically exact reflection layer. Our
method leaves every component of this inference graph unchanged.

\subsection{RDNet Backbone and Low-Pass Reflection Auxiliary Supervision}

\begin{figure*}[t]
    \centering
    \includegraphics[width=0.98\textwidth]{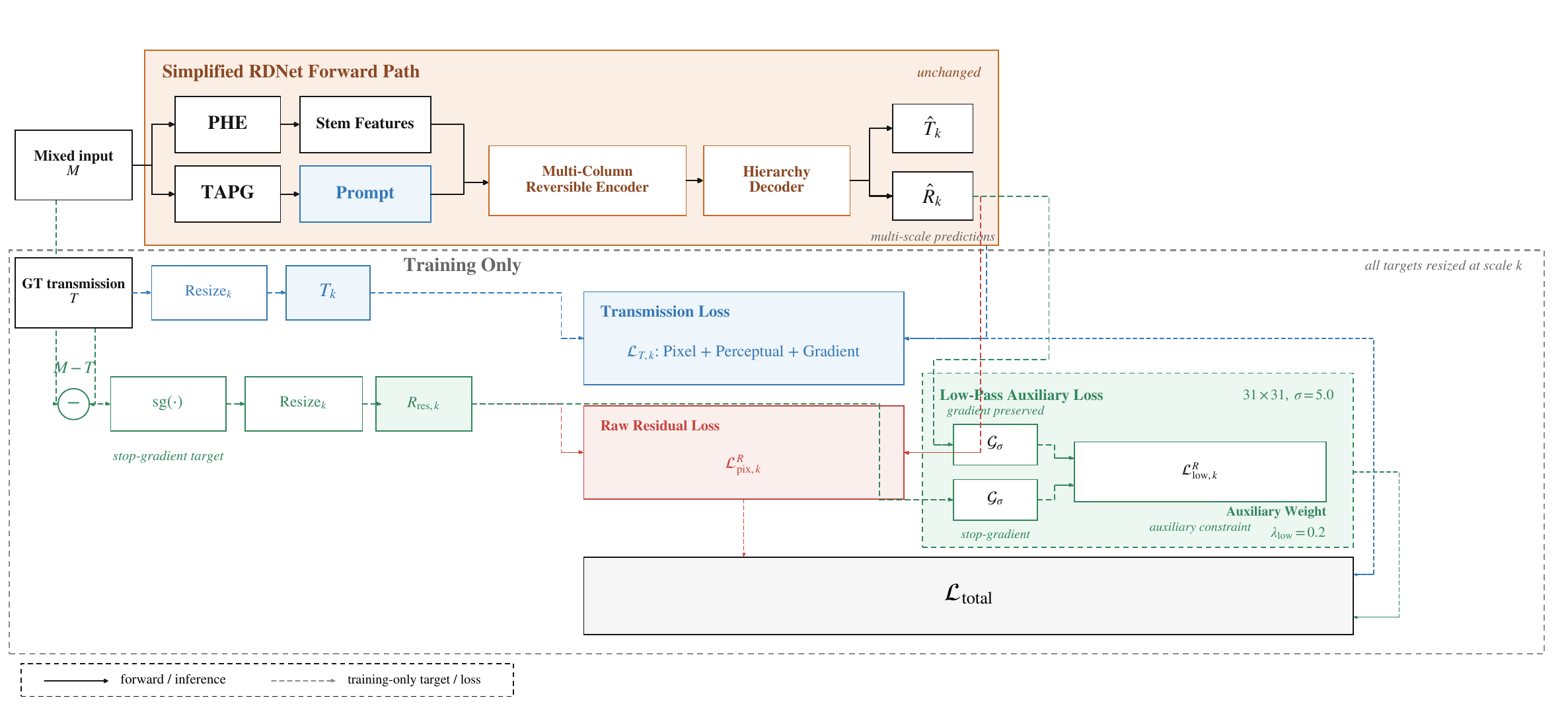}
    \caption{Proposed training supervision framework. The RDNet forward path is
    unchanged. At each output scale, the residual pseudo-label provides the
    main residual-branch loss, while symmetrically filtered prediction and
    pseudo-label provide an auxiliary low-pass constraint. Dashed components
    are used only during training.}
    \label{fig:training-framework}
\end{figure*}

Figure~\ref{fig:training-framework} summarizes the proposed training framework.
During training, the mixed image passes through RDNet once to produce
multi-scale $\hat T_k$ and $\hat R_k$. The clean transmission and residual
pseudo-label are resized independently for each output scale. For scale $k$,
the detached residual pseudo-label is
\begin{equation}
    R_{\mathrm{res},k} = \operatorname{Resize}_k\!\left(
        \operatorname{sg}(M-T)\right),
    \label{eq:scaled-target}
\end{equation}
where $\operatorname{sg}$ denotes stop-gradient. RDNet's original supervision
contains transmission pixel, perceptual, and gradient terms, together with a
residual-branch pixel loss:
\begin{align}
    \mathcal{L}_{T,k}
    &= \mathcal{L}^{T}_{\mathrm{pix},k}
     + \mathcal{L}^{T}_{\mathrm{per},k}
     + \mathcal{L}^{T}_{\mathrm{grad},k}, \\
    \mathcal{L}^{R}_{\mathrm{pix},k}
    &= \ell_{\mathrm{pix}}(\hat R_k,R_{\mathrm{res},k}),
    \label{eq:base-losses}
\end{align}
where $\ell_{\mathrm{pix}}$ is mean squared error.
The transmission terms serve complementary roles: pixel loss anchors color
and intensity, perceptual loss encourages feature-level similarity, and
gradient loss protects edges. The residual-branch pixel loss makes
decomposition explicit and discourages the transmission branch from
explaining all mixed content by itself.

In real pairs, $M-T$ can contain high-frequency components caused by small
registration errors, exposure changes, sensor noise, and the unmodeled term
$\Phi(T,R)$ in Eq.~\eqref{eq:formation-expanded}. Directly replacing the raw
target by a blurred target removes useful details, as verified by our
ablation. Instead, LowAux keeps the raw residual as the main supervision and
constrains only the low-frequency projection of $\Rres$ by applying the same
differentiable Gaussian operator $G_\sigma$ to prediction and target:
\begin{equation}
    \mathcal{L}^{R}_{\mathrm{low},k}
    =
    \ell_{\mathrm{pix}}\!\left(
       G_\sigma(\hat R_k), G_\sigma(R_{\mathrm{res},k})\right).
    \label{eq:lowpass-loss}
\end{equation}
We use a normalized depth-wise $31\times31$ kernel with $\sigma=5.0$ and
reflection padding. The prediction path remains differentiable; the target
path is detached. Consequently, the auxiliary loss constrains stable
low-frequency reflection structure without forcing $\hat R_k$ itself to be
blurred.

The symmetric filtering is important. Filtering only the target would ask an
unfiltered prediction to match a deliberately blurred image and would
directly suppress its high-frequency content. Filtering both sides instead
compares their low-frequency projections. Gradients still flow from
$G_\sigma(\hat R_k)$ into the RDNet prediction, while high-frequency details
remain governed by the raw residual loss. The complementary comparison between
R3 and R4 tests precisely this distinction.

The complete objective is
\begin{align}
    \mathcal{L}_{\mathrm{total}}
    &=
    \sum_k w_k\left(
        \mathcal{L}_{T,k}
        +\mathcal{L}^{R}_{\mathrm{pix},k}
        \right. \notag\\[-2pt]
    &\hspace{4.2em}\left.
        +\lambda_{\mathrm{low}}\mathcal{L}^{R}_{\mathrm{low},k}
    \right),
    \qquad \lambda_{\mathrm{low}}=0.2,
    \label{eq:total-loss}
\end{align}
where $w_k$ are RDNet's multi-scale weights. Thus, LowAux modifies only
optimization-time supervision while preserving RDNet's original inference
graph. It introduces no test-time parameters, memory, or computation.

\subsection{Scene-Balanced RRW Extension}

Low-pass auxiliary supervision improves the aggregate benchmark result but
does not fully address distribution shifts in Objects, Postcard, and Wild. We
therefore use real pairs from the published RRW dataset~\cite{zhu2024rrw}
while keeping the network and objective unchanged. Our contribution here is
the scene-balanced sampling strategy used to incorporate RRW, not the
dataset's collection or ground-truth acquisition procedure.

RRW contains many adjacent frames for each captured scene. Uniform frame
sampling would over-represent long videos and allow nearly identical frames to
dominate an epoch. We first match each scene with its scene-level ground truth,
discard scenes without a target, and split the 167 valid scenes into 147
training, 15 validation, and 5 holdout scenes using a fixed seed. Each epoch
samples exactly eight frames from every training scene, yielding 1,176
scene-balanced RRW pairs. Input and target receive identical crop, flip, and
rotation transforms, and the residual pseudo-label is constructed as
$\Rres=M-T$.
Splitting by scene, rather than by frame, prevents adjacent views of the same
capture from crossing between training and held-out partitions. Resampling
eight frames per scene at every epoch exposes the model to temporal variation
without allowing long sequences to dominate the gradient. This design treats
the scene as the statistical unit and makes RRW complementary to the existing
independent image sources.

\begin{table}[t]
    \centering
    \caption{Training mixture of the proposed method. The epoch size remains
    fixed at 7,932 samples.}
    \label{tab:training-mix}
    \small
    \begin{tabular}{lc}
        \toprule
        Training branch & Sampling ratio \\
        \midrule
        Real89 paired data & 15\% \\
        Nature200 paired data & 15\% \\
        VOC generic synthetic data & 55\% \\
        Scene-balanced RRW pairs & 15\% \\
        \bottomrule
    \end{tabular}
\end{table}

Table~\ref{tab:training-mix} shows the final mixture. RRW contributes real
diversity without replacing the existing paired and synthetic sources. It was
introduced after the low-pass-only model showed uneven behavior on Objects,
Postcard, and Wild, all of which contain real capture characteristics that are
poorly represented by generic synthetic mixtures. Public test sets are never
used as training sources, and validation and holdout RRW scenes never enter
the training sampler.

\begin{figure*}[t]
    \centering
    \begin{minipage}[t]{0.49\textwidth}
        \centering
        \includegraphics[width=\linewidth]{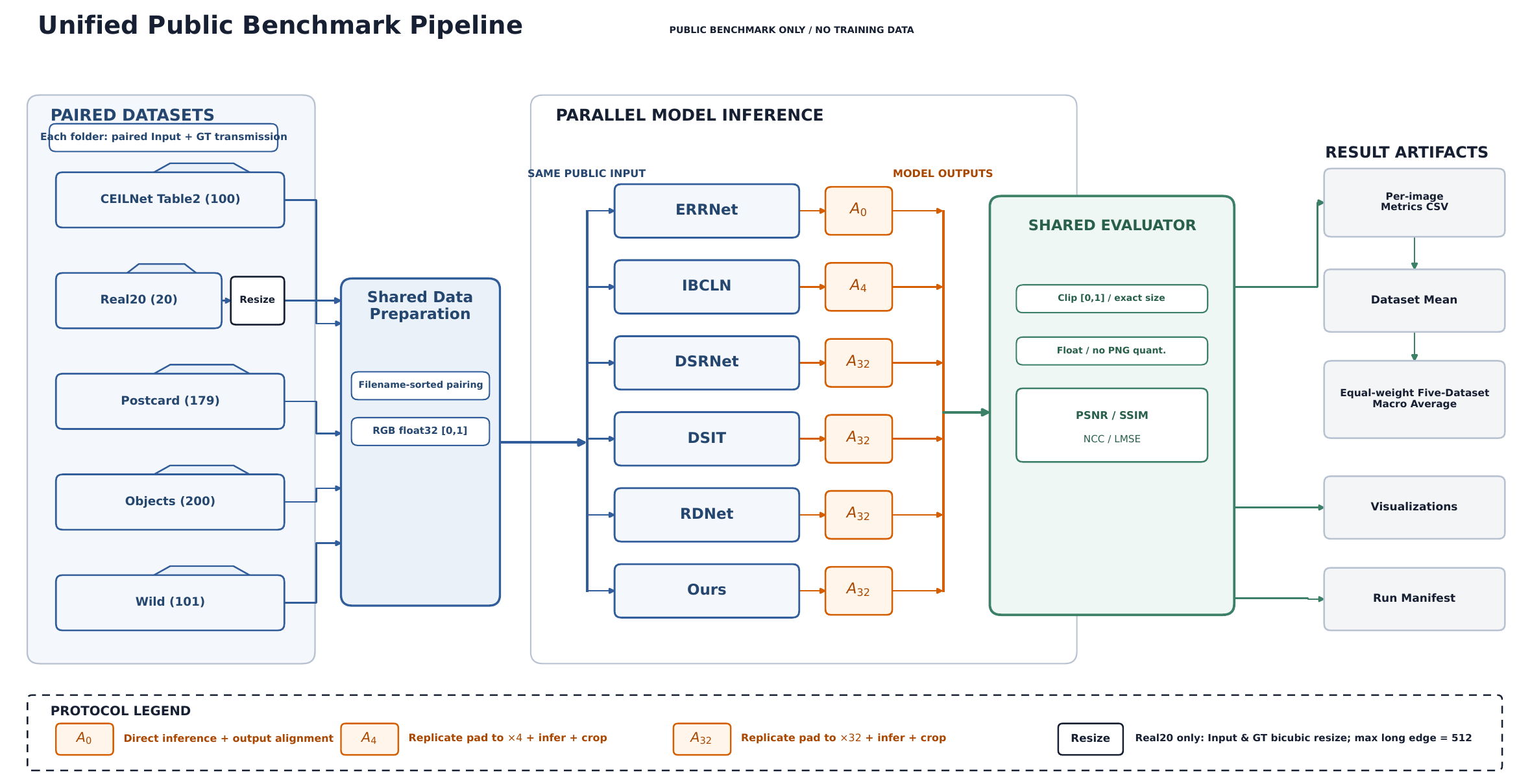}
        \captionof{figure}{Unified public benchmark. All models receive the
        same paired floating-point input and use the same evaluator. Adapters
        perform only required padding, inference, and cropping.}
        \label{fig:benchmark-pipeline}
    \end{minipage}
    \hfill
    \begin{minipage}[t]{0.49\textwidth}
        \centering
        \includegraphics[width=\linewidth]{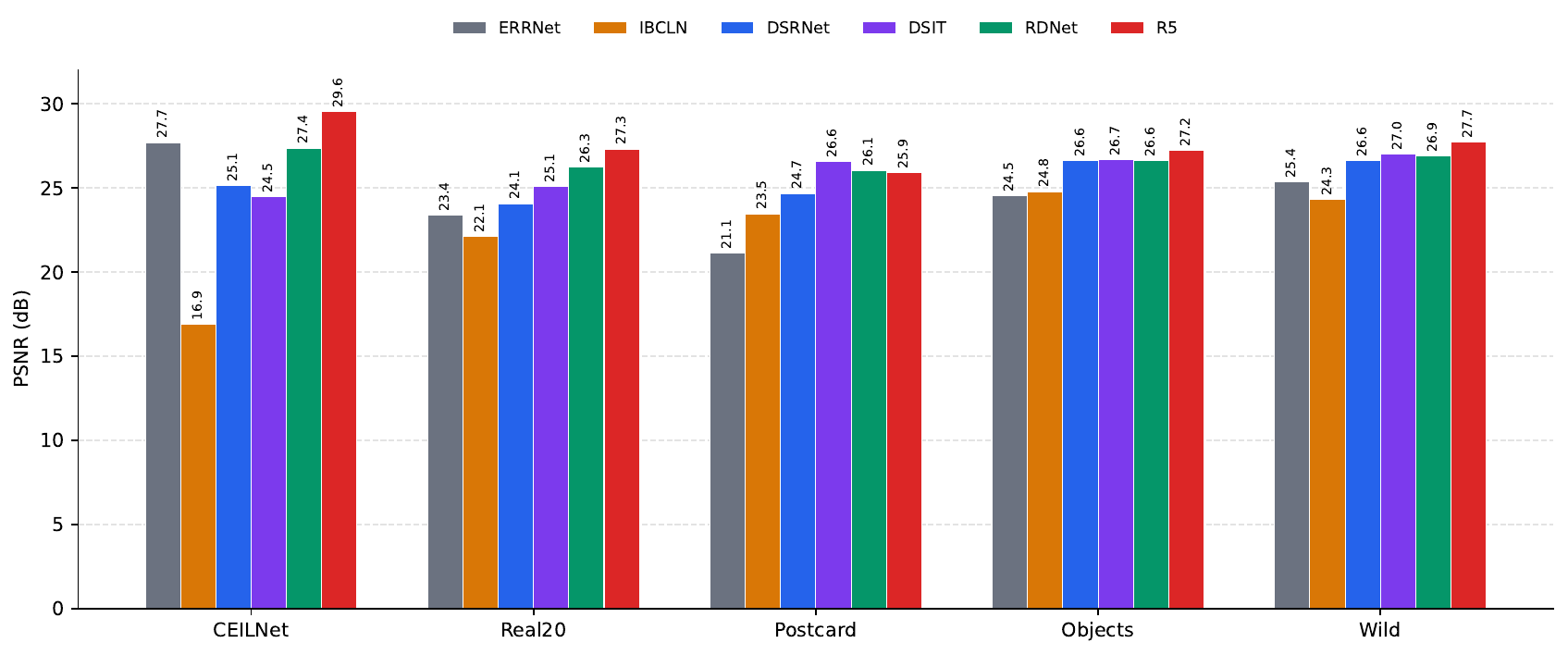}
        \captionof{figure}{Per-dataset PSNR under the unified benchmark.
        \ours is strongest among the compared models on four datasets, while
        DSIT performs best on Postcard.}
        \label{fig:dataset-psnr}
    \end{minipage}
\end{figure*}

\section{Experiments and Analysis}

\subsection{Unified Public Benchmark}

\paragraph{Protocol.}
The benchmark contains CEILNet Table2 (100 images), Real20 (20), Postcard
(179), Objects (200), and Wild (101), for 600 paired samples in total.
Inputs and ground truths are paired by identical sorted filenames and loaded
as RGB float32 tensors in $[0,1]$. For Real20 only, both input and target are
bicubically resized when the long edge exceeds 512 pixels. Model adapters
handle only mandatory inference differences: RDNet, DSRNet, and DSIT use
replicate padding to multiples of 32, while IBCLN uses padding to multiples of
four and its official three-round cascade. Every prediction must exactly match
the target size.

Predictions are clipped to $[0,1]$ and evaluated before PNG quantization using
one implementation of PSNR, SSIM, NCC, and LMSE. We first average images
within each dataset and then compute an equal-weight five-dataset macro
average, preventing larger subsets from dominating the main result.
Figure~\ref{fig:benchmark-pipeline} summarizes the protocol.

\paragraph{Metrics.}
PSNR measures absolute reconstruction fidelity and strongly penalizes
pixel-level error. SSIM evaluates local luminance, contrast, and structural
agreement, making it more sensitive to visible texture preservation. NCC
measures correlation between prediction and target and is useful when overall
intensity differs slightly, whereas LMSE emphasizes local squared error and
exposes spatially concentrated artifacts. Reporting all four avoids treating
a high PSNR alone as sufficient evidence of visually correct layer
separation.

\paragraph{Training details.}
The proposed system is optimized with AdamW using BF16 mixed precision on four
GPUs. The effective batch size is eight. Learning rates are
$5\times10^{-6}$ for the pretrained feature backbone and $10^{-5}$ for the
remaining parameters. Training uses the fixed mixture in
Table~\ref{tab:training-mix} and the objective in Eq.~\eqref{eq:total-loss}.
The final model is initialized from the low-pass-auxiliary RDNet checkpoint
and trained with RRW while retaining the same loss configuration.

\subsection{Comparison with Public Models}

\begin{figure*}[t]
    \centering
    \begin{minipage}{0.90\textwidth}
        \centering
        \addtocounter{table}{1}
        \captionof{table}{Five-dataset macro average under the unified
        public benchmark.}
        \label{tab:main-results}
        \small
        \setlength{\tabcolsep}{6.0pt}
        \renewcommand{\arraystretch}{1.08}
        \begin{tabular}{lcccc}
            \toprule
            Method & PSNR$\uparrow$ & SSIM$\uparrow$ & NCC$\uparrow$ & LMSE$\downarrow$ \\
            \midrule
            ERRNet~\cite{wei2019errnet} & 24.425 & 0.8844 & 0.9433 & 0.007842 \\
            IBCLN~\cite{li2020ibcln} & 22.313 & 0.8512 & 0.9231 & 0.010352 \\
            DSRNet~\cite{hu2023dsrnet} & 25.429 & 0.9000 & 0.9569 & 0.006847 \\
            DSIT~\cite{hu2024dsit} & 25.964 & 0.9085 & 0.9660 & 0.006002 \\
            RDNet~\cite{zhao2025rdnet} & 26.647 & 0.9133 & 0.9677 & 0.005253 \\
            \midrule
            \ours & \textbf{27.546} & \textbf{0.9220} & \textbf{0.9751} &
            \textbf{0.004760} \\
            \bottomrule
        \end{tabular}
    \end{minipage}
    \vspace{6pt}
    \includegraphics[width=0.99\textwidth]{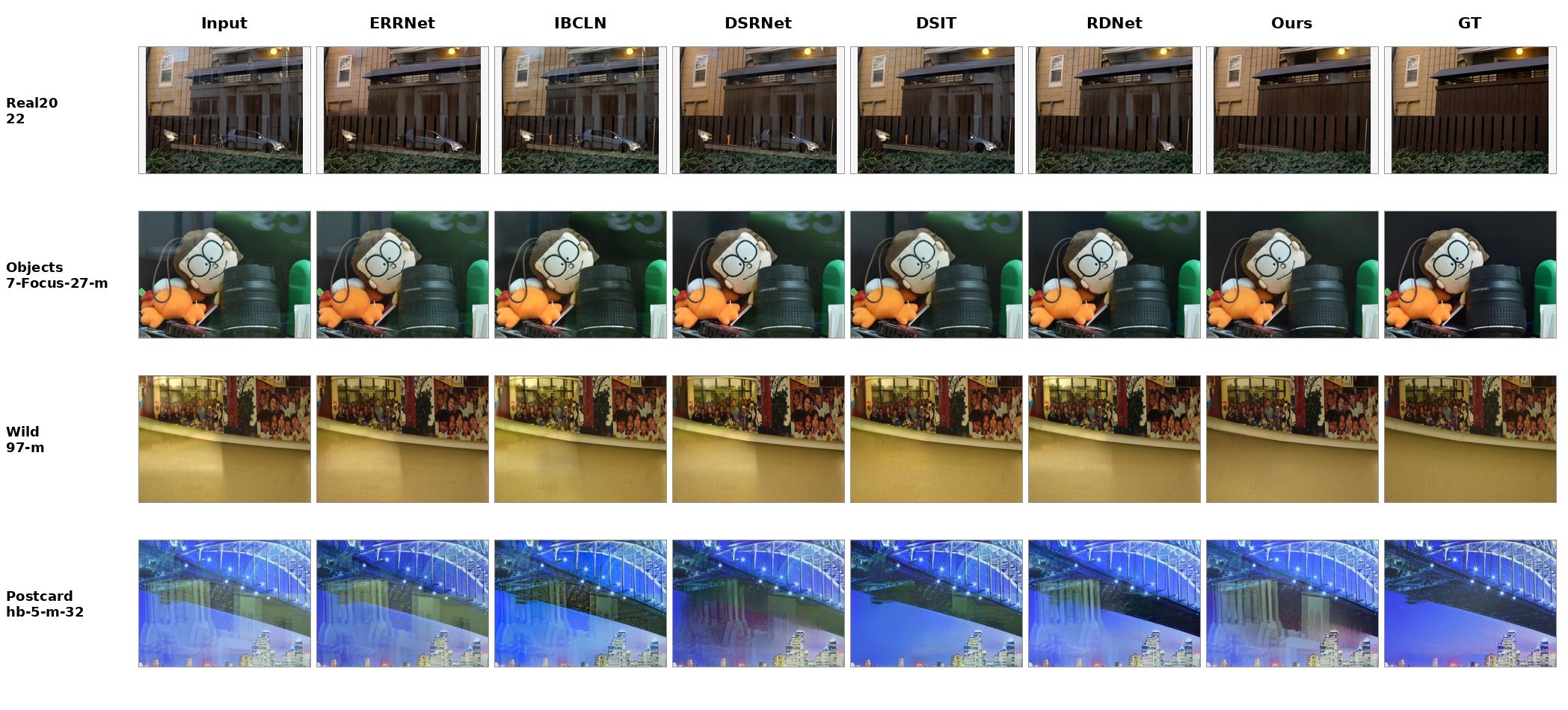}
    \caption{Qualitative comparison on representative public-benchmark
    samples. \ours removes the reflected car in Real20, the green reflected
    digits in Objects, and the triangular wall glare in Wild. The Postcard row
    exposes a failure to remove the pale reflected facade.}
    \label{fig:qualitative}
\end{figure*}

Table~\ref{tab:main-results} reports the common macro average. Under the
unified evaluation protocol, \ours achieves the strongest five-dataset
macro-average performance among the compared models across all four reported
metrics. Relative to ERRNet, it improves PSNR by
3.122 dB and SSIM by 0.0376, while reducing LMSE by 0.003081. It also exceeds
DSIT by 1.582 dB and the official RDNet checkpoint by 0.900 dB in macro PSNR.

Figure~\ref{fig:dataset-psnr} reveals substantial distribution-specific
behavior that is hidden by a single average. \ours reaches 29.565, 27.314,
27.227, and 27.709 dB on CEILNet, Real20, Objects, and Wild, respectively,
ranking first among the public-model comparison on these four datasets.
DSIT is strongest on Postcard at 26.555 dB, exceeding \ours by 0.641 dB.
This subset frequently contains two sharp, semantically coherent layers, so
successful removal requires deciding which complete structures belong to the
transmission rather than merely suppressing reflection-like frequencies.

\paragraph{Dataset-specific behavior.}
Figure~\ref{fig:dataset-psnr} makes the cross-dataset trade-offs explicit. On
CEILNet, \ours exceeds ERRNet and official RDNet by 1.86 and 2.19 dB,
respectively. This subset contains many regular synthetic or controlled
mixtures, for which explicit decomposition and the retained raw residual
constraint are particularly effective. On Real20, the 1.05 dB gain over
official RDNet indicates that the added supervision remains useful on real
paired images rather than only fitting synthetic residuals.

Objects and Wild emphasize more varied content, camera motion, reflection
strength, and spatially nonuniform artifacts. The results on these sets
support the use of scene-balanced RRW pairs: the model encounters a wider
range of real reflection appearances while preserving the original paired
and synthetic branches. Together with the Real20 gain, these results indicate
improved cross-dataset generalization on complex real scenes. Postcard is the
sole exception. Its sharp
overlapping semantic structures reward models that resolve layer ownership
correctly; frequency-aware supervision can stabilize residual learning but
cannot itself determine which complete object belongs to which layer.

\begin{table*}[t]
    \centering
    \caption{PSNR/SSIM on the four samples shown in
    Fig.~\ref{fig:qualitative}. Bold marks the best result for each sample.}
    \label{tab:sample-results}
    \scriptsize
    \setlength{\tabcolsep}{3.4pt}
    \begin{tabular}{lcccc}
        \toprule
        Sample & ERRNet & DSIT & RDNet official & \ours \\
        \midrule
        Real20/22 & 19.72 / 0.753 & 20.54 / 0.765 & 22.53 / 0.819 &
        \textbf{25.48 / 0.883} \\
        Objects/7-Focus-27-m & 21.13 / 0.837 & 23.42 / 0.882 &
        25.50 / 0.905 & \textbf{31.11 / 0.933} \\
        Wild/97-m & 22.31 / 0.953 & 23.16 / 0.957 & 21.16 / 0.941 &
        \textbf{29.50 / 0.969} \\
        Postcard/hb-5-m-32 & 20.26 / 0.822 & \textbf{25.36 / 0.886} &
        21.80 / 0.846 & 18.47 / 0.796 \\
        \bottomrule
    \end{tabular}
\end{table*}

Figure~\ref{fig:qualitative} provides an explicit visual comparison. In
Real20/22, most competing outputs retain the reflected silver car across the
lower-right fence. \ours removes almost all of the car while preserving the
vertical wooden slats and foreground ivy. In Objects/7-Focus-27-m, it removes
the large green reflected digits near the upper edge and the bright green
reflection at the right, yet preserves the plush toy's face and the fine ribs
of the camera lens. In Wild/97-m, it suppresses the triangular bright glare on
the yellow wall without blurring the edges of the photo collage above it.

The final Postcard example shows the opposite behavior. DSIT produces a
comparatively uniform blue region beneath the bridge and removes most of the
pale reflected facade. \ours retains the vertical translucent columns and a
reddish haze in the center. DSIT reaches 25.36 dB on this sample versus
18.47 dB for \ours, confirming that two sharp semantic layers remain a central
limitation.

The sample-level measurements in Table~\ref{tab:sample-results} agree with
the visible differences. On Objects/7-Focus-27-m, \ours gains 5.61 dB over
official RDNet; the recovered object boundaries remain sharp while the
high-contrast reflected structure is strongly attenuated. On Wild/97-m, the
7.34 dB advantage over DSIT accompanies cleaner flat regions without removing
the desired contours. Real20/22 shows a smaller but consistent improvement in
both reconstruction fidelity and structure. In contrast, the Postcard failure
reduces both PSNR and SSIM, indicating that the error is not merely a global
color shift: desired semantic structure is actually assigned to the wrong
layer.

The comparison is deliberately system-level. Public checkpoints were trained
with different data and optimization schedules, and \ours additionally uses
RRW. The unified evaluator removes test-time implementation discrepancies but
does not make the training conditions identical. Controlled evidence for our
design choices is therefore provided by the following RDNet method evolution
and complementary ablations.

\subsection{RDNet Method Evolution and Complementary Ablations}

\begin{table*}[t]
    \centering
    \caption{RDNet ablations under the unified benchmark. Macro PSNR uses the
    unified public benchmark unless marked otherwise.}
    \label{tab:rdnet-evolution}
    \small
    \setlength{\tabcolsep}{4.2pt}
    \renewcommand{\arraystretch}{1.10}
    \begin{tabular}{
        >{\raggedright\arraybackslash}p{0.18\textwidth}
        >{\raggedright\arraybackslash}p{0.34\textwidth}
        >{\centering\arraybackslash}p{0.12\textwidth}
        >{\raggedright\arraybackslash}p{0.22\textwidth}}
        \toprule
        Method & Modification & Macro PSNR$\uparrow$ & Conclusion \\
        \midrule
        RDNet reproduced & Reproduced RDNet training & 26.764 &
        Explicit-decomposition baseline \\
        R3 & Replace raw residual supervision with a low-pass residual target &
        26.821 & Low-frequency information alone is insufficient \\
        R4 LowAux & Add low-pass residual supervision as an auxiliary term &
        27.014 & More stable than replacing the raw residual target \\
        R5 RRW & R4 objective with scene-balanced RRW real pairs &
        \textbf{27.546} & Best aggregate result \\
        Sharp-Synth+RRW (archived) & Add sharp local synthetic reflections
        together with RRW & 26.252$^{*}$ & Degrades Postcard; not adopted \\
        \bottomrule
    \end{tabular}
    \vspace{2pt}

    \parbox{0.96\textwidth}{\scriptsize
    $^{*}$Sharp-Synth+RRW (archived) reports its epoch-4 short-run five-set macro PSNR,
    not a final unified public-benchmark result.}
\end{table*}

Table~\ref{tab:rdnet-evolution} summarizes the RDNet ablations. RDNet
reproduced establishes the explicit-decomposition baseline. R3 replaces the
full residual pseudo-label with a low-pass main target; it improves only
slightly over the reproduced baseline and remains below R4. This indicates
that low-frequency residual structure is more useful as an auxiliary
constraint than as a replacement for the raw residual target. R4 retains the
raw residual supervision and adds LowAux, reaching 27.014 PSNR, 0.91637 SSIM,
0.96951 NCC, and 0.005205 LMSE. R5 then adds scene-balanced RRW real pairs
without changing the loss, improving macro PSNR by a further 0.532 dB.
The archived Sharp-Synth+RRW variant degrades Postcard in short-run testing,
so the final pipeline retains RRW but removes the sharp-synthetic branch.

\subsection{Discussion}

\paragraph{Why the two improvements are complementary.}
The method evolution and complementary ablations indicate that supervision
design and data distribution address different failure modes. The complete
R4 LowAux-based training variant improves the aggregate result without
changing the training sources or deployed inference graph. The comparison
supports the usefulness of low-frequency auxiliary structure, while not
isolating it as the sole cause of the R4 gain. R5 then improves Real20,
Postcard, Objects, and Wild while retaining the same LowAux objective.
Because the objective is unchanged in this stage, the results support the
conclusion that scene-balanced RRW broadens real-scene coverage and improves
cross-dataset generalization, with clear gains on Real20 and Wild.

\paragraph{Macro balance versus image-weighted performance.}
Our primary score gives each dataset equal weight. This choice prevents the
200-image Objects subset and 179-image Postcard subset from dominating the
20-image Real20 subset, and it matches our objective of robust performance
across distributions. We also computed the image-weighted average over all 600
samples directly from the same dataset summaries. \ours remains first, with
27.309 PSNR, 0.9282 SSIM, 0.9804 NCC, and 0.003110 LMSE. Thus, the conclusion
is not an artifact of macro aggregation, although the numerical margin changes
with the weighting rule.

\paragraph{Fairness boundary.}
The public comparison guarantees identical test inputs, adapter constraints,
floating-point metrics, and aggregation, but it cannot equalize training data
or optimization. In particular, \ours uses additional RRW pairs, while the
public checkpoints follow their authors' original training recipes. The main
comparison therefore demonstrates the practical competitiveness of the final
system. Claims about individual design choices rely on the internal RDNet
method evolution and complementary ablations, where the backbone and evaluator
are held fixed whenever a quantitative comparison is made.

\paragraph{Practical cost.}
Both proposed changes are confined to training. Scene-balanced RRW sampling
changes which pairs enter an epoch, and Gaussian filtering is used only when
computing the reflection loss. The deployed model has the same parameters,
forward graph, padding rule, and output format as its RDNet backbone. This is
important for a practical restoration system: the gain does not require an
extra refinement network, an ensemble, or iterative inference.

\paragraph{Failure taxonomy.}
The qualitative examples reveal three recurring errors. Under-removal leaves
visible reflected edges in otherwise correct transmission; this often lowers
PSNR while retaining reasonable SSIM. Over-removal erases real transmission
texture and reduces both fidelity and local structure. The hardest case is
semantic layer confusion, where a sharp, coherent object is assigned to the
wrong layer. Such failures can look locally plausible yet disagree strongly
with the target, as in the selected Postcard example. Low-pass supervision
mainly addresses unstable residual structure and RRW broadens real-scene
coverage, but neither explicitly resolves semantic ownership.

\paragraph{Model selection and remaining limitation.}
The final checkpoint was chosen during exploratory development using the same
five-set macro protocol. This can introduce optimistic selection bias even
though no test images enter training. A stricter study should reserve an
independent validation set, select one checkpoint once, and evaluate it on an
untouched benchmark. Moreover, the persistent DSIT advantage on Postcard
shows that real-data diversity and frequency-aware supervision do not fully
solve semantic layer assignment. Explicit semantic consistency or
region-adaptive separation is a promising next step.

\section{Conclusion}

This work adopts RDNet as an explicit decomposition backbone and improves its
training through low-pass reflection auxiliary supervision and
scene-balanced RRW real pairs. Under the
unified floating-point benchmark, the resulting system achieves the highest
five-dataset macro-average PSNR, SSIM, and NCC and the lowest LMSE among the
compared models. The RDNet ablations show that low-pass information is more
effective as an auxiliary constraint than as a replacement target, while the
scene-balanced RRW extension improves cross-dataset balance. The archived
Sharp-Synth+RRW experiment is not adopted. Remaining Postcard failures
motivate stronger semantic layer-assignment cues.
\label{body:end}

\section*{Code Availability}

The source code, training scripts, evaluation pipeline, and experiment
materials are available at:

\begin{center}
\url{https://github.com/flyingc2004/lowaux-rdnet}
\end{center}

\appendix
\section{Supplementary Real-Scene Case Study}

In addition to the standard five-dataset benchmark, we include a small
supplementary evaluation on individually selected paired examples. The set
contains three self-captured pairs and two public real-world holdout samples
from RRW. The two RRW scenes,
\texttt{reflection\_hf2/outdoor\_wild5} and
\texttt{reflection\_vivo3/indoor\_office}, were kept outside the R5 training
and validation splits. This appendix set is used only as a case study and is
not included in the public five-dataset macro average reported in the main
paper.

\begin{figure*}[t]
    \centering
    \includegraphics[width=\textwidth]{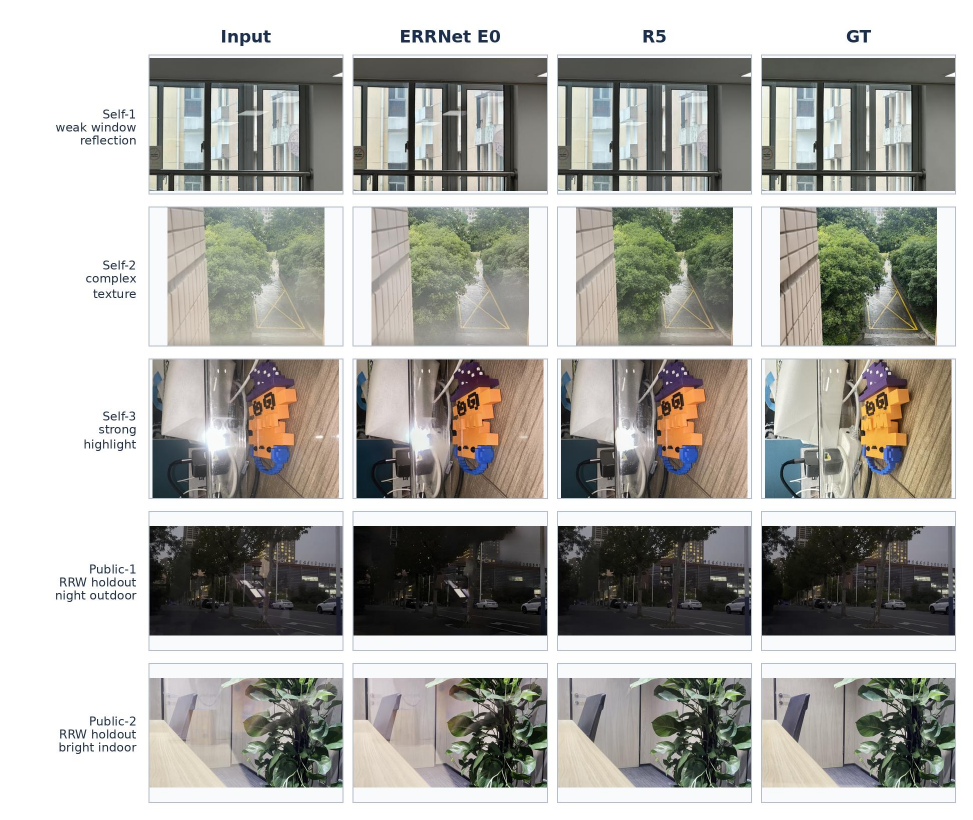}
    \caption{Supplementary five-image visual comparison. The first three rows
    are self-captured paired examples; the last two rows are RRW holdout
    real-world samples that were not used during R5 training or validation.
    Columns show the mixed input, ERRNet baseline output, R5 output, and
    reflection-free ground truth.}
    \label{fig:self-collected}
\end{figure*}

Table~\ref{tab:my5-all-models} reports the five-image average under the same
floating-point evaluator used by the public benchmark. R5 obtains the best
overall score on this small set, improving over the ERRNet baseline by
$+5.56$~dB PSNR, $+0.065$ SSIM, and $+0.080$ NCC, while reducing LMSE by
$0.0132$. The comparison in Fig.~\ref{fig:self-collected} and
Table~\ref{tab:my5-errnet-r5} further shows that the largest gains come from
the two held-out real scenes. For the weak window reflection case, R5 preserves
the transmission structure more stably. For the complex-texture case, R5
improves PSNR and NCC but slightly increases LMSE, indicating that local
regional errors remain. The strong-highlight example is the hardest case: R5
only brings a small numerical improvement, suggesting that saturated specular
reflections still exceed what the current residual supervision can reliably
separate.

\begin{table*}[t]
    \centering
    \caption{Five-image supplementary average. The set contains three
    self-captured pairs and two RRW holdout real-world samples.}
    \label{tab:my5-all-models}
    \small
    \begin{tabular}{lcccc}
        \toprule
        Method & PSNR$\uparrow$ & SSIM$\uparrow$ & NCC$\uparrow$ & LMSE$\downarrow$ \\
        \midrule
        ERRNet E0 & 18.38 & 0.691 & 0.803 & 0.0291 \\
        IBCLN & 21.49 & 0.730 & 0.872 & 0.0231 \\
        DSRNet & 20.44 & 0.722 & 0.800 & 0.0239 \\
        DSIT & 21.22 & 0.733 & 0.857 & 0.0211 \\
        RDNet official & 21.82 & 0.743 & 0.873 & 0.0203 \\
        \ours{} R5 & \textbf{23.94} & \textbf{0.756} & \textbf{0.882} & \textbf{0.0159} \\
        \bottomrule
    \end{tabular}
\end{table*}

\begin{table*}[t]
    \centering
    \caption{ERRNet baseline and R5 comparison on each supplementary example.}
    \label{tab:my5-errnet-r5}
    \scriptsize
    \resizebox{\textwidth}{!}{
    \begin{tabular}{llrrrrrrrrr}
        \toprule
        Sample & Scene type
        & \multicolumn{4}{c}{ERRNet E0}
        & \multicolumn{4}{c}{\ours{} R5}
        & $\Delta$PSNR \\
        \cmidrule(lr){3-6}\cmidrule(lr){7-10}
        & & PSNR & SSIM & NCC & LMSE & PSNR & SSIM & NCC & LMSE & \\
        \midrule
        Self-1 & Weak window reflection
        & 25.48 & 0.941 & 0.988 & 0.0028
        & 28.75 & 0.952 & 0.993 & 0.0016 & +3.27 \\
        Self-2 & Complex texture
        & 12.83 & 0.248 & 0.638 & 0.0434
        & 16.32 & 0.257 & 0.745 & 0.0486 & +3.49 \\
        Self-3 & Strong highlight
        & 13.80 & 0.680 & 0.695 & 0.0229
        & 14.16 & 0.691 & 0.702 & 0.0195 & +0.37 \\
        Public-1 & RRW holdout, night outdoor
        & 17.61 & 0.680 & 0.734 & 0.0703
        & 30.09 & 0.917 & 0.979 & 0.0087 & +12.48 \\
        Public-2 & RRW holdout, bright indoor
        & 22.16 & 0.907 & 0.958 & 0.0062
        & 30.38 & 0.966 & 0.994 & 0.0012 & +8.22 \\
        \bottomrule
\end{tabular}}
\end{table*}

\end{document}